\documentclass[runningheads]{llncs}
\usepackage{graphicx}
\usepackage{hyperref}
\usepackage{amsmath}

\begin{document}
\title{T2CI-GAN: Text to Compressed Image generation using Generative Adversarial Network}

\author{Bulla Rajesh\inst{1,2}\orcidID{0000-0002-5731-9755} \and
Nandakishore Dusa\inst{1} \and
Mohammed Javed\inst{1}\orcidID{0000-0002-3019-7401}\and
Shiv Ram Dubey\inst{1}\orcidID{0000-0002-4532-8996}\and
P. Nagabhushan \inst{1,2}}
\authorrunning{Bulla. Rajesh et al.}
%
\institute{Department of IT, IIIT Allahabad, Prayagraj, U.P, 211015, Idia \and Department of CSE, Vignan University, Guntur, A.P, 522213, India \\
\email{\{rsi2018007, iwm2016002, javed, srdubey, pnagabhushan\}@iiita.ac.in}}

\maketitle              
\begin{abstract}
The problem of generating textual descriptions for the visual data has gained research attention in the recent years. In contrast to that the problem of generating visual data from textual descriptions is still very challenging, because it requires the combination of both Natural Language Processing (NLP) and Computer Vision techniques. The existing methods utilize the Generative Adversarial Networks (GANs) and generate the uncompressed images from textual description. However, in practice, most of the visual data are processed and transmitted in the compressed representation. Hence, the proposed work attempts to generate the visual data directly in the compressed representation form using Deep Convolutional GANs (DCGANs) to achieve the storage and computational efficiency. We propose GAN models for compressed image generation from text. The first model is directly trained with JPEG compressed DCT images (compressed domain) to generate the compressed images from text descriptions. The second model is trained with RGB images (pixel domain) to generate JPEG compressed DCT representation from text descriptions. The proposed models are tested on an open source benchmark dataset Oxford-102 Flower images using both RGB and JPEG compressed versions, and accomplished the state-of-the-art performance in the JPEG compressed domain.
The code will be publicly released at GitHub after acceptance of paper.
\keywords{Compressed Domain \and Deep Learning \and DCT Coefficients \and T2CI-GAN \and JPEG Compression \and Compressed Domain Pattern Recognition \and Text to Compressed Image.
}
\end{abstract}
\section{Introduction}
\label{sec:intro}
Generating visually realistic images based on the natural text descriptions is an interesting research problem that warrants knowledge of both language processing and computer vision. Unlike the problem of image captioning that generates text descriptions from image, the challenge here is to generate semantically suitable images based on proper understanding of the text descriptions. Many interesting techniques have been proposed in the literature to explore the problem of generating pixel images from the given input texts \cite{text2image}, \cite{stackGAN}, \cite{SD-GAN}, \cite{mirrorGAN}. Moreover, a very recent attempt by \cite{compressedImage} is aimed to generate images in the compressed format. The whole idea here is to avoid synthesis of RGB images and subsequent compression stage. In fact, in the current digital scenario, more and more images and image frames (videos) are being stored and transmitted in compressed representation. The compressed data in the internet world has reached more than 90\% \cite{bulla2020} of traffic. On the other hand, different compressed domain technologies are being explored both by the software giants, like Uber \cite{gueguen2018faster} and Xerox \cite{de1998fast}, and academia \cite{javed2018review}, \cite{mukhopadhyay2011image}, \cite{tompkins1999fast}, \cite{bell2001pattern}, that can directly process and analyse compressed data without decompression and re-compression. Some of the prominent works in compressed document images are discussed in \cite{javed2013extraction,javed2014extraction,javed2015direct} and \cite{bulla2020,bullacict}. This gives us strong motivation for exploring the idea of generating compressed images directly from natural text descriptions, and that is attempted in this research paper.

\begin{figure}[!t]
\centering
\includegraphics[scale=0.30]{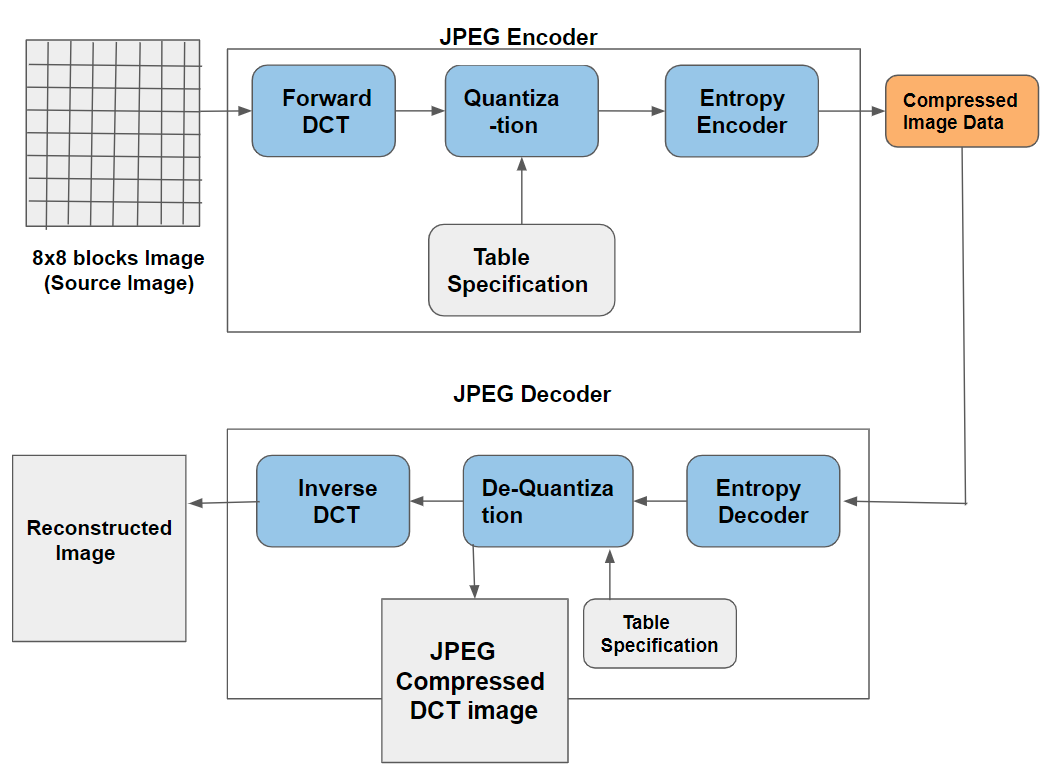}\par
\caption{JPEG Compression and Decompression architecture and extraction of JPEG Compressed DCT image which is used in the proposed approach.}
\label{4}
\vspace{-15pt}
\end{figure}

 Recently, Generative Adversarial Network (GAN) models have been successfully used for generating realistic images from diverse inputs such as layouts \cite{he2021context}, texts\cite{xu2018attngan}, and scenes \cite{ashual2019specifying}. However, early GAN models \cite{text2image} have generated images of low resolutions from the input text. In \cite{text2image}, the GAN model was used to generate image from a single sentence. This method was implemented in two stages. Initially the text sentence was encoded into a feature matrix using deep CNNs and RNNs to extract the significant features. Then those features were utilized to generate a picture. In order to improve the quality, a stacked GAN was reported in \cite{stackGAN}. It generated the output picture using two GANs. In the first step, GAN-1 produced a low resolution image with basic shape and colors along with the background generated from a random noise vector. In the second step, GAN-2 improvised the produced image by adding details and making some required corrections. 
MirrorGAN was reported in \cite{mirrorGAN} for text to image translation through re-description. This model has reported the improved semantic consistency between text and produced output image. 
In \cite{SD-GAN}, authors proposed a Semantics Disentangling Generative Adversarial Network (SD-GAN) which exploited the semantics of text description. 
However, all the GAN based techniques discussed above were trained using RGB pixel images meant to generate RGB images. Hence, our work is focused on employing the significant features of GAN for generating compressed images directly from the given text descriptions.

In the recent literature, a GAN model was proposed for generating direct compressed images from noise vector \cite{compressedImage}. Since JPEG compression was the most used format, the authors attempted to generate direct JPEG compressed images rather than generating RGB images and compressing them separately. Their GAN framework consists of Generator, Decoder and Discriminator sub networks. The Generator consists of locally connected layers, quantization layers, and chroma subsampling layers. These locally connected layers perform the block based operations similar to JPEG compression methods to generate JPEG compressed images. In between the Generator and the Discriminator, a Decoder was used to decompress the image to facilitate the comparison with ground truth RGB image by the Discriminator network. In specific, this decoder performed de-quantization and Inverse Discrete Cosine Transformation (IDCT) followed by YCbCr to RGB transformations on the compressed images generated by the Generator. Unlike \cite{compressedImage} which generates the compressed images from noise, our model generates the compressed images based on the given input text descriptions.

Overall, this research paper propose two novel GAN models for generating compressed images from text descriptions. The first GAN model is trained directly with JPEG compressed DCT images to generate compressed images from text description. The second GAN model is trained with RGB images to generate compressed images from text descriptions. The proposed models have been tested on Oxford-102 Flower images benchmark dataset using both the RGB and JPEG compressed versions, reporting state-of-the-art performance in the compressed domain. Rest of the paper is organized as follows: Section II presents the preliminaries of used concepts. Section III discusses the proposed methodology and GAN architectures. Section IV reports the detailed experimental results and analysis. Finally, Section V concludes the paper with a summary.

\section{Preliminaries}
In this section, a brief description of JPEG compression, GAN model and GloVe model is presented.


\subsection{JPEG Compression}
JPEG compression algorithm achieves compression by discarding the high frequency components. Firstly, the RGB channels of the image are converted into YCbCr format to separate the luminance (Y) and chrominance (CbCr) channels as,
\vspace{-7pt}
\begin{equation}
Y = (0.299 \times r + 0.587 \times g + 0.114 \times b) 
\end{equation}
\vspace{-12pt}
\begin{equation}
   Cb = (- 0.1687 \times r - 0.3313 \times g + 0.5 \times b + 128)
\end{equation}
\vspace{-10pt}
\begin{equation}
   Cr = (0.5 \times r - 0.4187 \times g - 0.0813 \times b + 128).
\end{equation}
Then each channel is divided into $8 \times8$ non-overlapping pixel blocks. Forward Discrete Cosine Transform (DCT) is applied on each block in each channel to convert the $8 \times8$ pixel block (let's say $P(x,y)$) from spatial domain to frequency domain. Each DCT block, i.e.,  $F(u,v)$, is quantized to keep only the low frequency coefficients. Then Differential pulse code modulation (DPCM) is applied on the DC components and Run Length Encoding (RLE) on AC components. Huffman Coding is used to encode the DC and AC components in smaller number of bits. In order to perform the decompression, Entropy decoding, De-Quantization, and Inverse DCT (IDCT) are applied in the given order on the compressed image to obtain the uncompressed image. The compression and decompression stages are illustrated in Fig. \ref{4}. 
In the proposed work, the JPEG compressed DCT images are directly extracted from the JPEG compressed stream and used for training the deep learning model. The decompression is done only for the performance analysis, otherwise it is not required in practice. 

\subsection{Generative Adversarial Network (GAN)}
Generative Adversarial Network (GAN) \cite{goodfellow2014generative} is a deep learning model built with two networks, including Generator and Discriminator. The Generator (G) generates new images in the training images distribution and the Discriminator (D) classifies the images between actual and generated images into real and fake categories, respectively. 
These two sub models are trained alternatively such that Generator(G) tries to fool the Discriminator by generating data similar to real domain, whereas the Discriminator is optimized to distinguish the generated images from the real images. Overall, the Generator and the Discriminator play a two player min-max game. The objective function of the GAN is given as follows:
\begin{multline}
    \label{eq4}
    \min_{G}\max_{D} F(G,D) = E_{y \sim k_{d}}[\log D(x)] + 
     \\ E_{z \sim k_{z}}[\log (1-D(G(z))]
\end{multline}
where $y$ indicates real image sampled from $k_{d}$ (true data distribution), $z$ indicates noise vector sampled from $k_{z}$ (uniform or Gaussian distribution).

\begin{figure}[!t]
\begin{center}
		\includegraphics[ scale=.31]{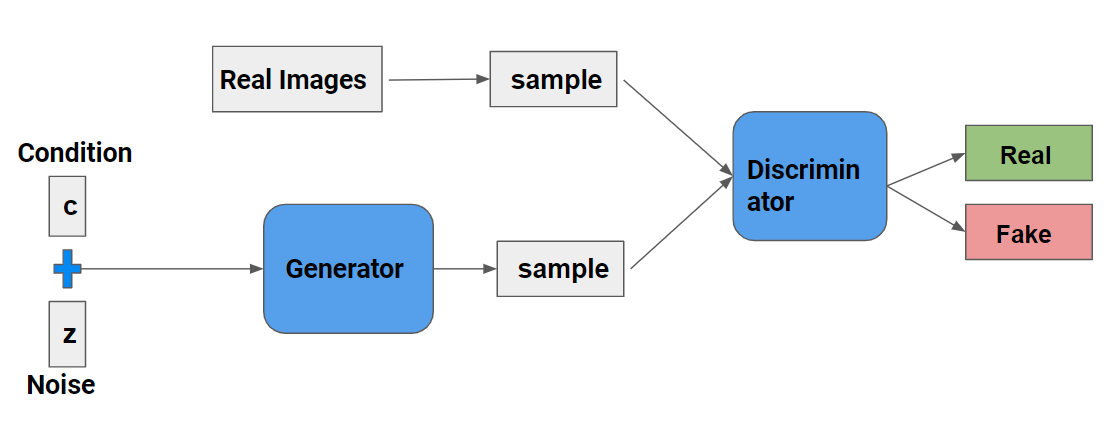}\par
			\vspace{-15pt}
\caption{Conditional GAN architecture \cite{mirza2014conditional}.}
			\label{2}
\end{center}
\vspace{-15pt}
\end{figure}

The Conditional GAN model \cite{mirza2014conditional} makes use of some additional information along with the noise. Both Generator (G) and Discriminator (D) use this additional information which is referred as  conditioning variable `c' that can be text or any other data. Thus, the Generator on Conditional GAN generates the images conditioned on variable `c' as depicted in Fig.~\ref{2}.

\subsection{GloVe Model}
GloVe stands for Global Vectors \cite{GloVe}. GloVe is an unsupervised learning algorithm. It is used for obtaining vector representations of words. It is an Open Source project developed at stanford. Word vectors make words having same meaning to cluster together and dissimilar words to repel. Word2Vec depends on local context information of words. The advantage of GloVe is that, unlike Word2vec to exctract or produce word vectors, it incorporates global statistics.

\begin{figure*}[!t]
\begin{center}
\includegraphics[scale=0.40]{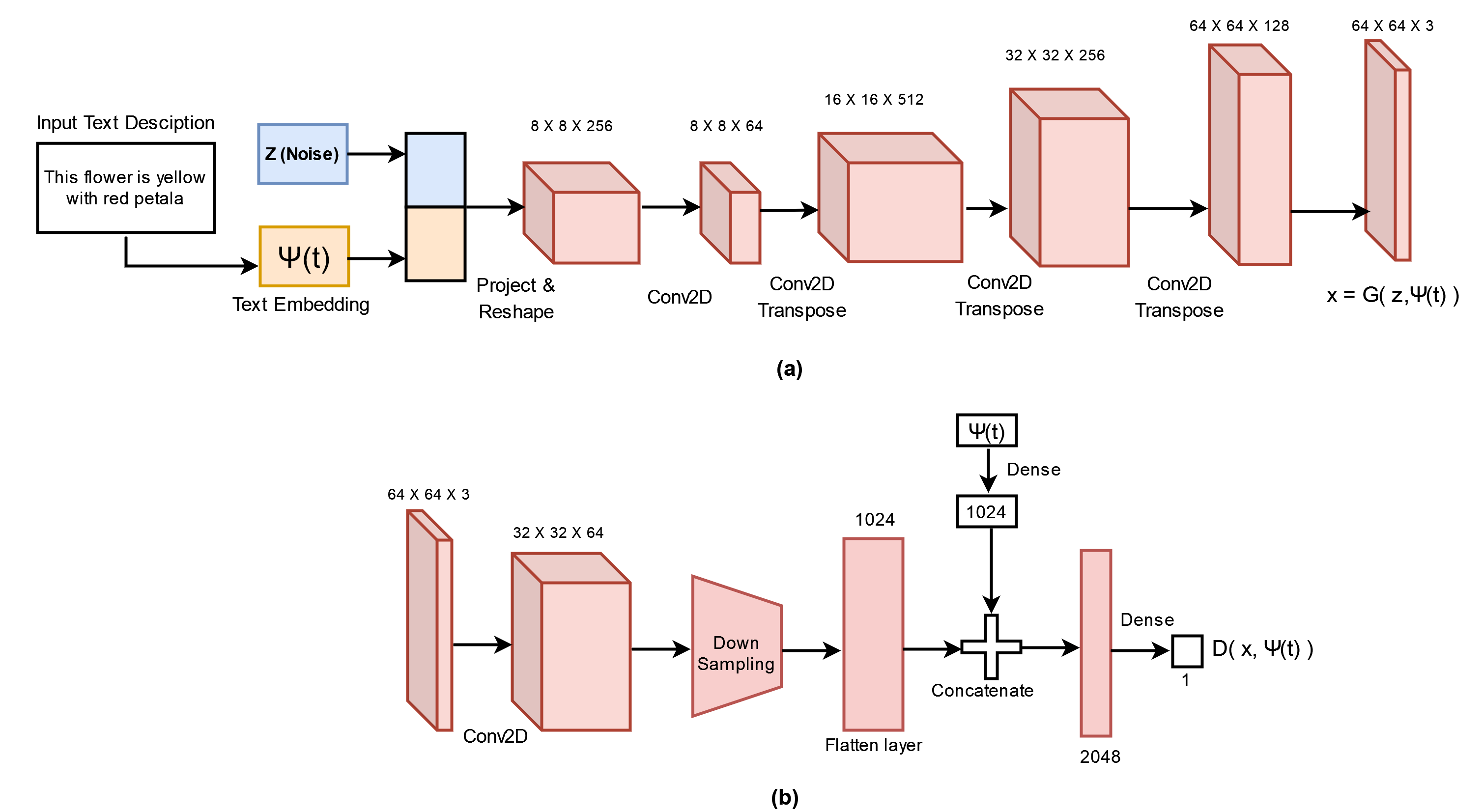}
\caption{The proposed T2CI-GAN Model-1 architecture using the backbone networks of \cite{text2image}. (a) Generator network and (b) Discriminator network.}
\label{3}
\end{center}
\vspace{-15pt}
\end{figure*} 

\section{Proposed Methodology}
In this section, we present the details of the proposed T2CI-GAN models for text to compressed image translation. We propose two variants of T2CI-GAN with simple Generator and customized Generator, respectively. First, we discuss the base network architecture of the proposed T2CI-GAN models which is adapted from T2I-GAN \cite{text2image}.

\subsection{Network Architecture}
The Deep Convolutional GAN (DC-GAN) architecture shown in Fig.~\ref{3} is used for the implementation of text to image synthesis. This is our base architecture (T2CI-GAN) for implementation of text to compressed image generation. First, the Word Embeddings $\psi(t)$ of given text-descriptions are obtained using the GloVe model. It is concatenated with the Noise Vector `z' and used as input to the Generator. Hence we provide a pair of both Text Embedding vectors $\psi$(t) and noise (z) as the input to T2CI-GAN model, instead of only noise.

\subsubsection{Generator Network}
The embedding of size 300 and the noise of size 100 are concatenated and given as input to the Dense layer. After Reshaping the output of the Dense layer, a series of Convolutions and Batch Normalization are performed, respectively for 4 times. It is then followed by Convolution2D Transpose and finally a Convolution with `Tanh' activation function. The `LeakyReLU' activation function is used after every Batch Normalization. It produces a image tensor of shape $64\times64\times3$ which is given as input to the Discriminator.
\vspace{-5pt}
\subsubsection{Discriminator}
Discriminator takes the output of Generator ($64\times64\times3$ image tensor) and word embedding of size 300 as input. A series of 2D strided Convolutions are performed on image tensor. Batch Normalization is applied after all Convolutions except for first one. Before applying last Convolution, embedding is concatenated with previous output. The `LeakyReLU' activation function is used and followed by `Dropout' layer to all Convolution layers except for last layer where `Sigmoid' is used.

\subsection{Proposed T2CI-GAN Model-1: Training with JPEG Compressed DCT Images}
\label{method1}
\subsubsection{Preparing JPEG Compressed DCT Image dataset}
\label{transformation}
This is an important step for the proposed model. The JPEG compressed entropy encoded images of the dataset are partially decompressed to obtain DCT Compressed version by applying entropy decoding and De-Quantization steps in the decoder as shown in Fig.~\ref{4}. Sample images from the dataset in RGB and JPEG Compressed DCT image formats are shown in Fig.~\ref{5} and Fig.~\ref{7}, respectively. Note that after conversion of RGB image into JPEG compressed DCT form, the images will be in the form of coefficients upon which a custom transformation is applied  where a specified coefficients are selected to decrease the computational complexity and ease the training process. In this transformation, since JPEG is applied block wise ($8\times8$ blocks) on an image, first 5 coefficients from first row, first 3 coefficients from the second row, and the first coefficient from the third row are extracted from every $8\times8$ block and making other values to zero. From Fig.~\ref{8} shown, we see very slight decrease in image quality compared to the one before transformation.

\subsubsection{Normalization of the JPEG Compressed DCT Image}
It is important to normalize the DCT coefficients extracted from above paragraph for training the model. Unlike the original RGB dataset whose pixel values range from 0 to 255, the exact range of values in the compressed image are not known. So, maximum and minimum values are computed from all the DCT values of all images in the dataset. Then, using this maximum and minimum values, the DCT pixel values in the range from [-1, +1] are generated.

\subsubsection{Loss Function}
Binary Cross-Entropy loss function is used in binary classification tasks. It is also known as log loss \cite{logloss} and given as,
\begin{equation}
    BCE\_Loss = (-\dfrac{1}{N}) \sum_{i=1}^{N}  x_{i}(\log p(x_{i})) + (1-x_{i})(\log 1-p(x_{i}))
\end{equation}
where $x_{i}$ represents the actual class and $\log p(x_{i})$ is the probability of that class and N is the total number of instances.

\subsection{Training T2CI-GAN Model-1}
Noise vector of size 100 concatenated with word-embedding vector of size 300 is given as input to the Generator network which performs Up-samplings and Convolutions, and produces a $64\times64\times3$ image as output. This image-tensor is passed to Discriminator to classify it as fake (generated) or real (original). Training a GAN is a very challenging task, because both Generator and Discriminator networks are trained simultaneously. The main goal of GAN training is to find a point of equilibrium between the Generator and Discriminator models. So, it makes training of GAN unstable. For stable training \cite{Jason2019}, Batch-normalization is used in both Generator and Discriminator networks. The `LeakyReLU' activation function is used in all layers of Generator and Discriminator except for output layer. The `tanh' activation function is used for last layer in case of Generator and the `Sigmoid' activation function in case of Discriminator. 

\begin{figure*}[!t]
\begin{center}
	\includegraphics[scale=.55]{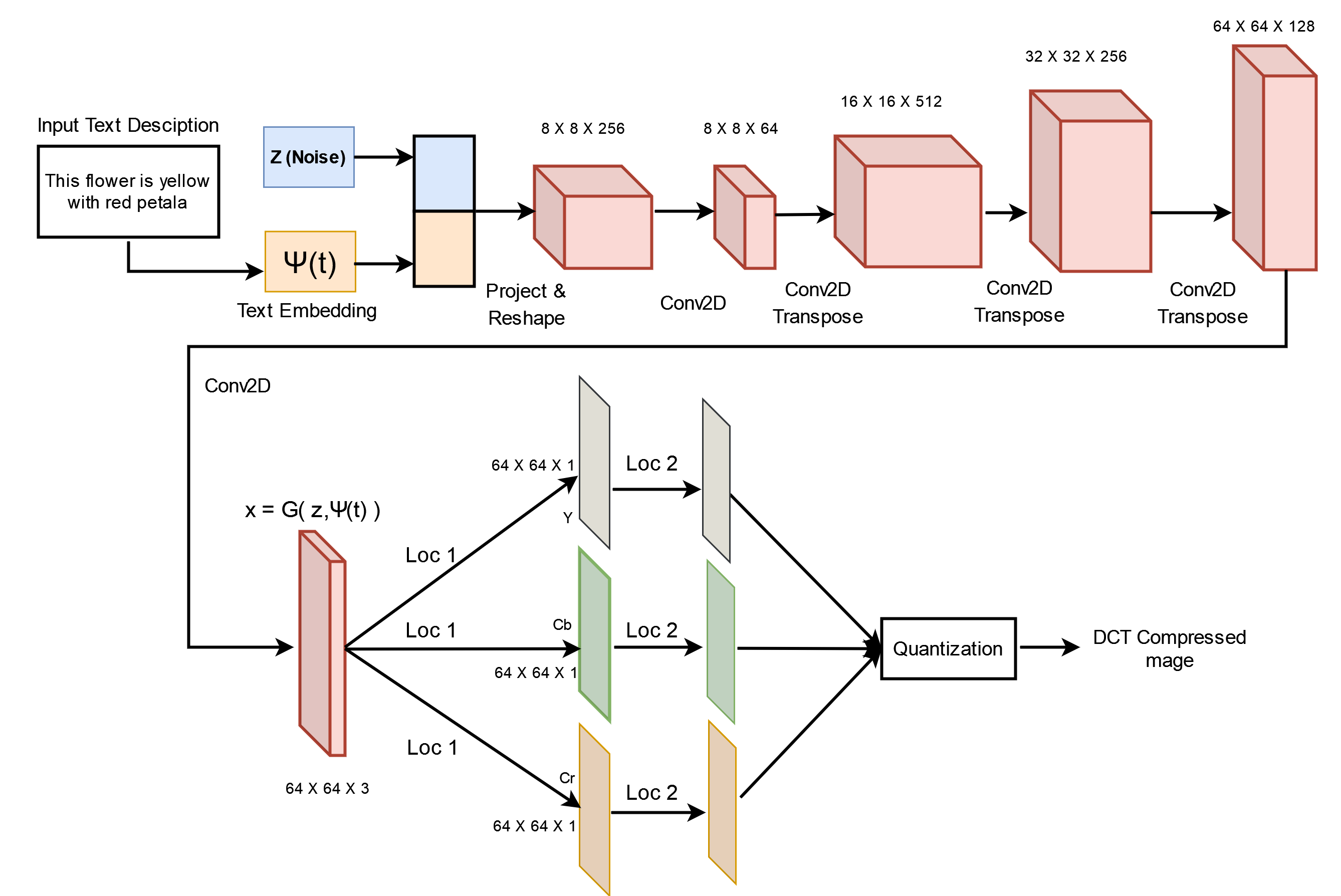}
	\caption{The Generator network of the proposed T2CI-GAN Model-2. The Discriminator is same as used in T2CI-GAN Model-1.)}
	\label{10}
\end{center}
\vspace{-15pt}
\end{figure*}

For training the proposed GAN Model-1, GAN-INT method \cite{text2image} is used, where the Discriminator is prepared on three pairs of inputs (original images, original captions), (generated images, original captions), and (original images, wrong captions). Binary Cross Entropy (BCE) loss function with ADAM optimizer is used for both Generator and Discriminator networks with learning rate = $0.0002$ and momentum = $0.5$. The BCE loss of generator is measured by first decoding the output of generator into RGB channels.   
The network is trained for 500 epochs with a mini-batch size of 64. After training the Discriminator network is discarded. For calculation of Generator loss, right descriptions is passed to the model as input. Then, loss is computed on the outputs from Discriminator (ranges between 0 and 1) and tensor of 1's. For calculation of Discriminator loss, three pairs of inputs (original images, original captions) for real loss, (generated images, original captions) and (original images, wrong captions) for fake loss are considered. The sum of the real loss and fake loss is regarded as the Discriminator loss. Overall, the T2CI-GAN model-1 uses same architecture as used by T2I-GAN Fig.~\ref{3}, but trained with JPEG compressed DCT images.

The proposed model is able to generate good JPEG Compressed DCT images close to the ones in the dataset, but on decompression to the RGB domain, the images either get distorted or suffers with the mode collapse problem by generating the similar images as shown in Fig.~\ref{14}. We conclude that the T2CI-GAN Model-1 fails to learn the RGB image information from JPEG Compressed DCT images. Hence, we propose T2CI-GAN Model-2 next which solves this problem.

\subsection{Proposed T2CI-GAN Model-2 : With Modified Generator and Training with RGB Images}
\label{method2}
It is noticed in T2CI-GAN Model-1 and in \cite{compressedImage} that while training on compressed images directly, DCT values fluctuate to a greater extent within the blocks and also across the blocks. Thus, the Discriminator gives sub-par quality gradients to the Generator and makes it difficult to train a decent Generator. Therefore, in this backdrop, we propose T2CI-GAN Model-2 for text to compressed image translation with modified Generator having the manual compression and decode modules as utilized by \cite{compressedImage} for compressed image generation from noise. The Discriminator network in T2CI-GAN Model-2 is same as in T2CI-GAN Model-1.

\subsubsection{Modified Generator}
In this proposed model, the Generator is modified such that it implements the idea of a JPEG encoder as used by \cite{compressedImage}. Basically, we develop the Generator of T2CI-GAN Model-2 by adding 6 locally connected layers along with a quantization layer and an entropy encoder layer in the Generator of T2CI-GAN Model-2. The modified Generator is shown in Fig.~\ref{10}.
After the Generator generates a $64\times64\times3$ image tensor from text and noise, the image tensor is divided into 3 channels, and each channel is passed through a Locally connected layer (Loc1) to generate Y, Cb, and Cr channels, respectively. Block size $1\times1$ is used in Loc1 layer. The Y, Cb, and Cr channels are then passed through another Locally connected layer (Loc2) which performs $8\times8$ block-wise operation like DCT. Basically, the Loc2 layer produces amplitudes of DCT for Y, Cb, and Cr channels, respectively. Next, the quantization is performed using the standard JPEG quantization method, where DCT block values are divided by a quantization matrix (based on quality factor). Finally an entropy encoder layer is used after the training.

\begin{figure}[!t]
        \center
		\includegraphics[ scale=0.3]{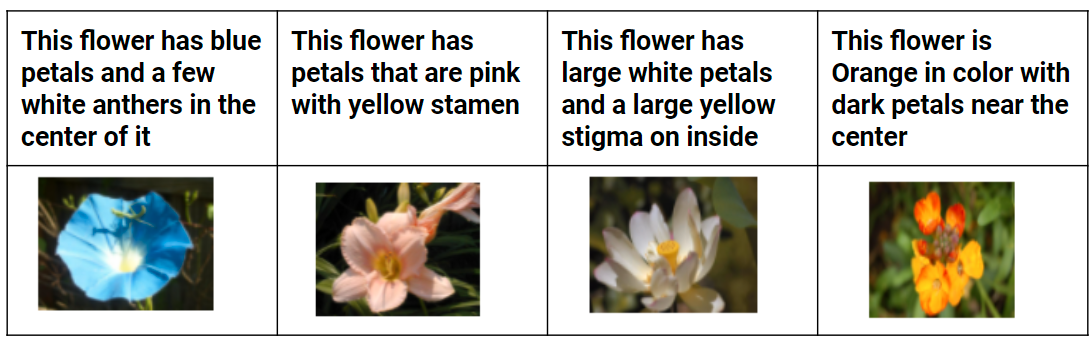}
			\caption{Sample RGB images from Oxford 102 Flowers dataset with input text descriptions.}
			\label{5}
			\vspace{-3pt}
\end{figure}

\begin{figure}[!t]
        \center
		\includegraphics[ scale=0.3]{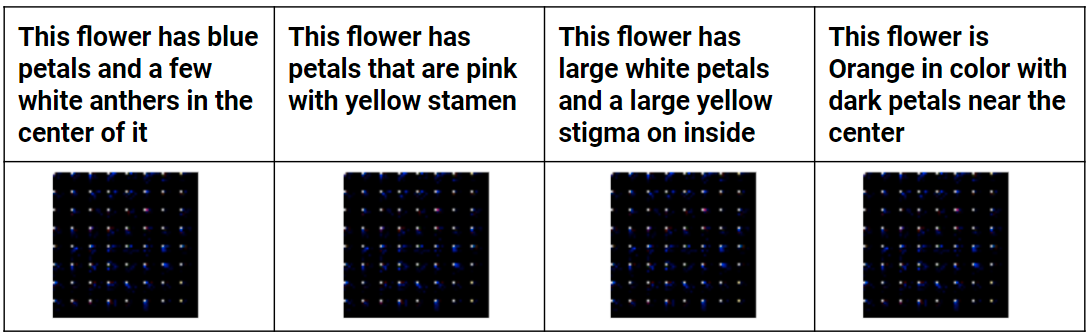}
			\caption{Corresponding sample JPEG Compressed DCT images generated from Oxford-102 Flower images shown in Fig.~\ref{5} and input text descriptions.}
			\label{7}
			\vspace{-15pt}
\end{figure}

\subsubsection{Decoder}
In the proposed T2CI-GAN Model-2, the Generator network generates the compressed versions of images (upto quantization) while trained using RGB image dataset. So, during training it is needed to decompress the compressed images into the RGB domain. For this, we use a non-trainable decoder ($H$) which takes input from the Generator and converts it into RGB pixel values and transfers the output to the Discriminator. In the decoder, De-quantization, Inverse Discrete Cosine Transform (2D IDCT) and color transformation are performed from YCbCr to RGB. At last, pixel values are clipped to range [0,255], and this is given as input to the Discriminator network.

\subsubsection{Loss Function}
The loss function remains same for Discriminator, but for Generator extra loss term $\gamma|H(G(z,t)) - $\^{G}$ (z,t)|$ is added to guide the locally connected layers similar to \cite{compressedImage}. Here $H$ is the decoder, $G$ is Generator, $z$ is noise vector, \^{G} indicates the layers of Generator before any Locally connected layer has been used, and
$\gamma$ is a hyperparameter as weight between original Generator loss and the modified Generator loss. The value of $\gamma$ used in the experiments is 0.1 

\section{Experimental Results}
\subsection{Oxford-102 Flowers Dataset}
\begin{figure}[!t]
        \center
		\includegraphics[ scale=0.40]{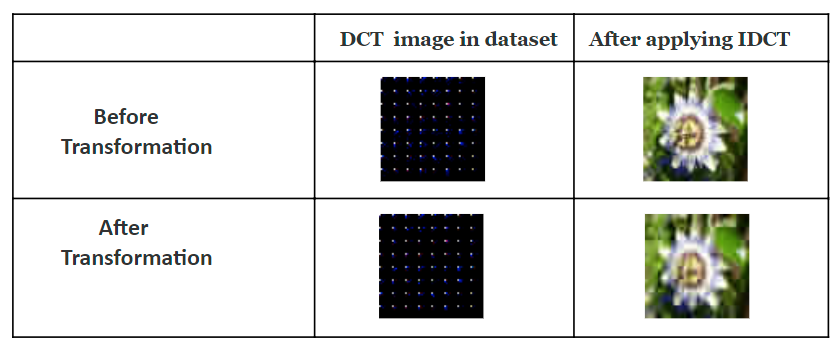}\par
			\caption{The significance of applying the simple transformation on JPEG Compressed DCT images. }
			\label{8}
			\vspace{-15pt}
\end{figure}

Oxford-102 flowers dataset \cite{Oxford102} has 102 flower categories in RGB format. Each class contains around 40 to 258 pictures with total 8189 images. Each image of the dataset has around 10 text-descriptions or captions portraying that image. To train the proposed models 2500 flower images are JPEG compressed to generate JPEG compressed Oxford-102 flowers dataset. Some sample images and captions of 102-flowers dataset are shown in Fig.~\ref{5}. The corresponding sample captions with JPEG Compressed Oxford-102 flowers dataset are shown in Fig. \ref{7}. The above transformation is applied on JPEG Compressed flowers dataset to ease the training process. Fig. \ref{8} shows sample flower image before and after transformation.

\subsection{Text to Compressed Image Results}
\subsubsection{T2CI-GAN Model-1 Results}
\begin{figure}[!t]
        \center
 		\includegraphics[ scale=0.43]{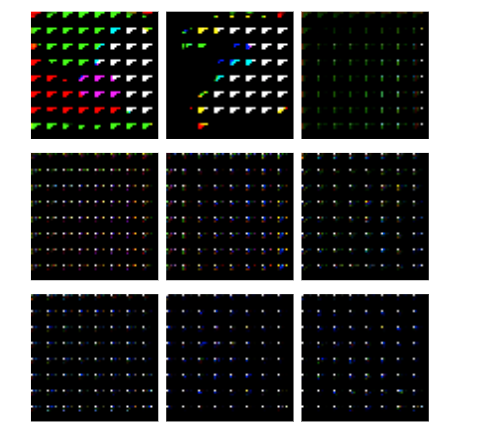}
			\caption{Sample output images generated for the text descriptions during training with T2CI-GAN Model-1.}
			\label{13}
			\vspace{-5pt}
\end{figure}

\begin{figure}[!t]
        \center
 		\includegraphics[scale=0.29]{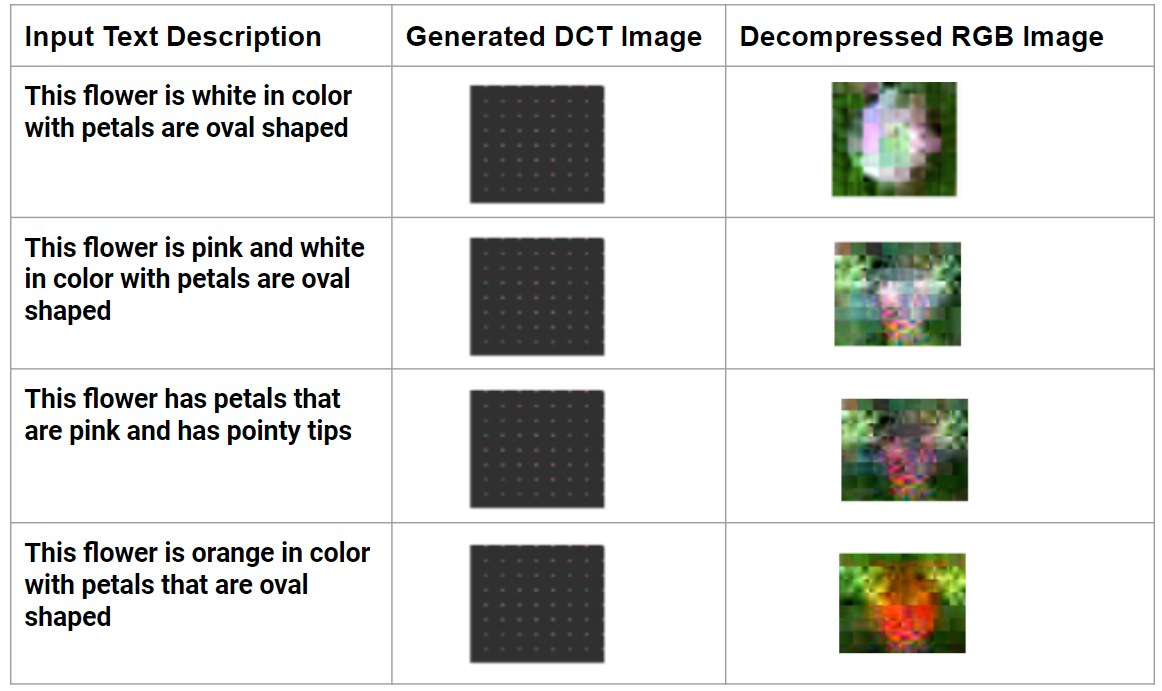}
			\caption{Sample output images generated for the text descriptions during training with the T2CI-GAN Model-1.}
			\label{14}
\end{figure}

The T2CI-GAN Model-1 is trained with 2500 JPEG Compressed DCT Oxford-102 flowers dataset and corresponding captions. The model is trained for 500 epochs. Fig.~\ref{13} shows the generated compressed images during training of the model. As mentioned earlier, the quality of generated DCT images is good, but after decompressing them to RGB format, the resultant images are either distorted or the same kind of images are generated for different input captions. 
Fig. \ref{14} shows different input texts, their corresponding generated JPEG compressed images, and their corresponding decompressed RGB images. It can be seen that only in a few cases the model gives correct results irrespective of the generated compressed images which look very similar for different input texts.

\begin{figure}[!t]
\center
\includegraphics[width=0.49\columnwidth, trim={0 7.84cm 0 0},clip]{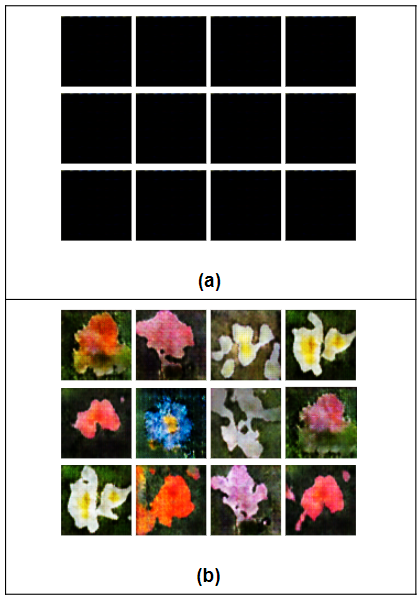}
\includegraphics[width=0.49\columnwidth, trim={0 0 0 7.84cm},clip]{snapshot3.png}
\caption{Performance of T2CI-GAN Model-2, where (a) sample JPEG compressed DCT images generated during training and (b) their corresponding decompressed images in the RGB domain.}
\label{14_a}
\end{figure}

\begin{figure}
        \center
 		\includegraphics[scale=0.35]{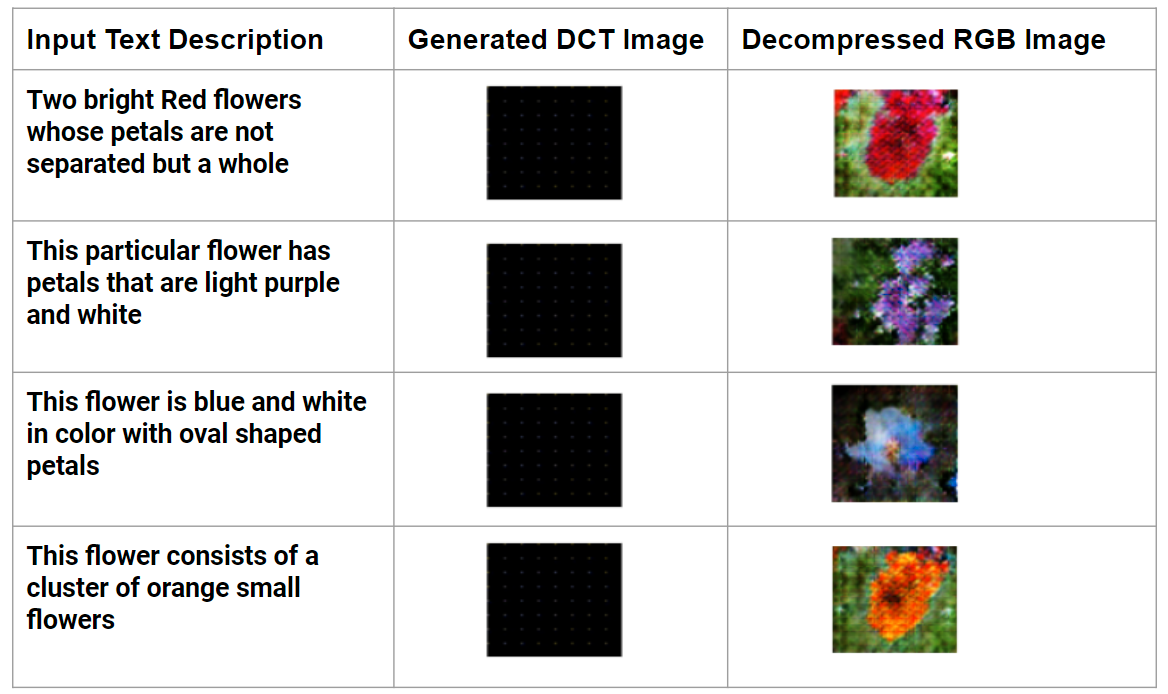}
		\caption{Sample output images generated for the input text descriptions with the T2CI-GAN Model-2.}
		\label{15}
\end{figure} 

\subsubsection{T2CI-GAN Model-2 Results}
In T2CI-GAN Model-2, the decoder is used to convert the compressed version of images generated by the Generator back into RGB format to feed them to the Discriminator network during training. This model is trained for 500 epochs. Fig. \ref{14_a} shows the compressed images (Quantized DCT images) generated and corresponding decoded RGB images during training. Fig. \ref{15} shows the sample Input text descriptions, Generated compressed version of image (Quantized DCT image) and its corresponding decoded RGB image. It can be observed that quality of the generated and decompressed images using T2CI-GAN Model-2 is much better than quality of the generated and decompressed images using T2CI-GAN Model-1.

\subsection{Quantitative Evaluation of the Model}
In order to perform the quantitative evaluation, We compute the InceptionScore \cite{Inceptionscore} for the proposed models. Inception score is a metric which evaluates the quality of generated images. It is mostly used for GANs evaluation which uses a pre-trained InceptionV3 model \cite{InceptionV3}. Using this model we calculate conditional probability of each generated image $p(y|x))$ and average of these conditional probabilities which is marginal probability $p(y)$. Next, we calculate KL divergence for each generated image as follows:
\begin{equation}
    KL = p(y|x) \times (log(p(y|x)) - log(p(y)).
\end{equation}
Summation of KL divergence over all images and average over all classes and exponent of this result is inception score.
Inception score captures quality of images and image diversity (i.e., whether a wide range of images are generated or not).

Table \ref{table1} shows the experimental of T2I-GAN \cite{text2image}, T2CI-GAN Model-1 and T2CI-GAN Model-2, where input domain for training of the model, domain of Generated images, domain of images to calculate inception score and inception scores are shown. Table \ref{table2} shows inception scores of methods proposed for text to compressed image generation. Here, instead of calculating Inception score directly with generated images in compressed domain, we first decompress them into RGB domain and then calculate the Inception score. From the results, it is clear that T2CI-GAN Model-2 performs better compared to T2CI-GAN Model-1 and achieves promising and state-of-the-art results in the compressed domain.

\begin{table}[!t]
\scriptsize
\begin{center}
    \caption{Performance comparison of T2I-GAN, T2CI-GAN Model-1 and T2CI-GAN Model-2 on Oxford-102 Flower dataset images with output and Inception Score measured in their respective domains.}
    \begin{tabular}{p{0.11\columnwidth}p{0.167\columnwidth}p{0.217\columnwidth}p{0.125\columnwidth}p{0.185\columnwidth}}
    \hline
    \textbf{Model} & \textbf{Input Domain} & \textbf{Generator Output} & \textbf{Domain} & \textbf{Inception Score} \\[1ex]
    \hline
    T2I-GAN & RGB  & RGB & RGB & 2.38 $\pm$ .17  \\
    Model-1 & Compressed & Compressed & Compressed & 1.08 $\pm$ .01 \\ 
    Model-2 & RGB & Compressed & Compressed & 1.01 $\pm$ .01  \\
    \hline
    \end{tabular}
    \label{table1}
\end{center}
\end{table}

\begin{table}[!t]
\scriptsize
\begin{center}
    \caption{Performance of the proposed compressed domain models, i.e., T2CI-GAN Model-1 and T2CI-GAN Model-2, on Oxford-102 Flower dataset with output in compressed domain and Inception Score measured in RGB domain. }
    \begin{tabular}{p{0.11\columnwidth}p{0.167\columnwidth}p{0.217\columnwidth}p{0.105\columnwidth}p{0.185\columnwidth}}
    \hline
    \textbf{Model} & \textbf{Input Domain} & \textbf{Generator Output} & \textbf{Domain} & \textbf{Inception Score} \\[1ex]
    \hline
    Model-1 & Compressed & Compressed & RGB & 1.42 $\pm$ .02 \\
    Model-2 & RGB & Compressed  & RGB & 2.01 $\pm$ .12  \\
    \hline
    \end{tabular}
    \label{table2}
\end{center}
\vspace{-15pt}
\end{table}

\begin{table}[!t]
\scriptsize
\begin{center}
    \caption{Comparison of the proposed compressed domain models with the state-of-the-art models such as GAN-INT-CLS \cite{text2image}, StackGAN \cite{stackGAN} and T2I-GAN \cite{text2image} using inception score measured in RGB domain on Oxford-102 dataset.} 
    \begin{tabular}{p{0.24\columnwidth}p{0.167\columnwidth}p{0.217\columnwidth}p{0.2\columnwidth}}
    \hline
    \textbf{Model} & \textbf{Input Domain} & \textbf{Generator Output} & \textbf{Inception Score} \\[1ex]
    \hline
    GAN-INT-CLS \cite{text2image} & RGB & RGB & 2.66 $\pm$ .03 \\
    StackGAN \cite{stackGAN} & RGB & RGB & 3.20 $\pm$ .01 \\
    T2I-GAN \cite{text2image} & RGB  & RGB  & 2.38 $\pm$ .17 \\
    T2CI-GAN Model-1 & Compressed & Compressed & 1.42 $\pm$ .02 \\
    T2CI-GAN Model-2 & RGB & Compressed & 2.01 $\pm$ .12 \\
    \hline
    \end{tabular}
    \label{table3}
\end{center}
\vspace{-15pt}
\end{table}

\subsection{Comparative Study}
The proposed T2CI-GAN models translate the text descriptions into compressed domain images. Although there is no existing GAN framework that generates compressed images directly from the input text, to compare our proposed models with state-of-the-art text to image GAN models, we first decompress the generated compressed images to RGB domain, and then evaluate and compare them with state-of-the-art models using Inception score. In Table \ref{table3}, we can see that inception score of the proposed models are low when compared with the state-of-the-art models like GAN-INT-CLS \cite{text2image}, StackGAN \cite{stackGAN} and T2I-GAN \cite{text2image} working in RGB domain because Inception score is usually calculated with 20k-50k images in the existing models. However, we train the model only with 2500 images and the corresponding captions to decrease the training time. So the Inception score reported with the models is low.

\section{Conclusion}
In this paper, we achieve the objective of generating compressed versions of images directly from text descriptions instead of generating raw RGB images and compressing it later as a post-processing step. We present two T2CI-GAN frameworks that generate compressed versions of images. We demonstrate the training and testing on Oxford flower-102 dataset. We observe a promising performance by the proposed T2CI-GAN Model-2. The performance of the proposed T2CI-GAN Model-1 is also satisfactory in compressed domain, but suffers to get the qualitative uncompressed images.  The proposed models can be further improved by using the state-of-the-art text to image GAN models and optimization techniques to improve the quality of generated images in the compressed domain.

 \bibliographystyle{splncs04}
 \bibliography{samplepaper}

\end{document}